# Robustly Pre-trained Neural Model for Direct Temporal Relation Extraction

Hong Guan,[1] Jianfu Li,[2] Hua Xu,[2] Murthy Devarakonda[1]

[1]Arizona State University

[2]The University of Texas Health Science Center

## Abstract

<u>Background:</u> Identifying relationships between clinical events and temporal expressions is a key challenge in meaningfully analyzing clinical text for use in advanced AI applications. While previous studies exist, the state-of-the-art performance has significant room for improvement.

<u>Methods:</u> We studied several variants of BERT (Bidirectional Encoder Representations using Transformers) some involving clinical domain customization and the others involving improved architecture and/or training strategies. We evaluated these methods using a direct temporal relations dataset which is a semantically focused subset of the 2012 i2b2 temporal relations challenge dataset.

<u>Results:</u> Our results show that RoBERTa, which employs better pre-training strategies including using 10x larger corpus, has improved overall F measure by 0.0864 absolute score (on the 1.00 scale) and thus reducing the error rate by 24% relative to the previous state-of-the-art performance achieved with an SVM (support vector machine) model.

<u>Conclusion:</u> Modern contextual language modeling neural networks, pre-trained on a large corpus, achieve impressive performance even on highly-nuanced clinical temporal relation tasks.

## Introduction

Identifying relationships between clinical events and temporal expressions is crucial for extracting insights from clinical narratives and for using the insights in advanced AI applications. Automatic means of extracting temporal information from clinical narratives has recently gained attention from the community.[1–7] While several types of temporal information can be extracted, in this paper we study relations between events and temporal expressions, which are critical for building timelines of clinical events. As shown in Figure 1, the task here is to identify the relation between the event "a magnetic resonance imaging" and a temporal expression "October 18, 1996" as Overlap, given the event and the temporal expression.

Semantic complexity of a task can cloud our ability to understand the difficulty and develop effective methods. Intuitively, the most common temporal relation

He had <u>a magnetic resonance imaging</u> performed on <u>October 18, 1996</u>.

Event: an magnetic resonance imaging
Temporal Expression: October 18, 1996
Relation type: Overlap

*Figure 1. An example of temporal relation extraction studied in this paper.*

is between an event and a temporal expression as in the example of Figure 1. However, a temporal relation may also exist between two events such as in "worsening of a condition after admission to hospital", and between two temporal expressions as in "two days before the date of admission". In the 2012 i2b2 challenge[2] all such relations were considered. However, Lee et al[8] identified the complexity of such a task and simplified the task and the dataset without compromising the difficulty, coverage, and importance of the task. We therefore, consider this important subset of temporal relations, which Lee et al call *direct* temporal relations.

A direct temporal relation is a relation between a temporal expression and an event that exists within a limited syntactic distance. Furthermore, in a direct temporal relation, a temporal expression modifies an event or vice versa, or the temporal expression and the event are arguments or adjuncts of the same predicate in a parse tree. The dataset with such direct temporal relations was created using a systematic approach including verification by medical experts. Three types of relations, as in the original 2012 i2b2 dataset, were included: Before, After, and Overlap. The task studied in this paper is to identify the three direct temporal relations in the Lee et al dataset.

Previously, different methods have been proposed to extract temporal relations from clinical notes, including machine learning techniques such as SVM,[6,9,10] Markov Logic Network (MLN),[11] structured learning,[12] and temporal constraints based document level model (TimeText).[13,14] In addition to the i2b2 2012 temporal relations challenge,[1] SemEval Clinical TempEval challenges from 2015 to 2017 tackled increasingly complex tasks starting from temporal information extraction to cross-domain temporal relation extraction.[3–5]

Nevertheless, recently developed deep learning technologies, such as pre-trained context-based language models, have not been extensively investigated for clinical temporal relation extraction. For example, the BERT (Bidirectional Encoder Representations using Transformers) neural network model, proposed in late 2018, has established new state-of-the-art results for several general domain[15] and clinical NLP tasks.[16–18] Recently several variations of the BERT model have emerged each with its own advantages and limitations. Variations of BERT were also created using clinical text and biomedical scientific articles for pre-training the model instead of the general domain text. Taking advantage of these recent developments, we studied six different BERT variations, namely the original BERT itself, RoBERTa,[19] ALBERT,[20] XLNet,[21] BioBERT,[22] and ClinicalBERT,[23] for extracting direct temporal relations. Our results establish new state-of-the-art performance for direct temporal relations with 0.7579 precision, 0.6932 recall, and an F measure of 0.7241, which is an 0.0864 absolute F measure improvement (on the 1.00 scale) over the previous best[8] (an SVM model) and a resulting error reduction of 24%.

## Methods

### Dataset

The direct temporal relation corpus, as the original 2012 i2b2 corpus, consisted of 310 discharge summaries that were split into the training set of 190 documents and the test set of 120 documents. Lee et al[8] developed direct temporal relations from the original i2b2 corpus in three steps. First, the transitive closure of all relations was determined. Second, only the intra-sentential relations between temporal expressions and events were kept. Finally, each of the selected relations was verified by domain experts as a direct temporal relation. (See Lee et al[8] for details of the process and the inter-annotator agreement).

Even though the 2012 i2b2 challenge initially used a more granular relation types, the challenge eventually used only "Overlap", "Before", and "After". The direct temporal relation dataset also used the three relations and the type distribution was very similar in both datasets. Table 1 shows the type distribution of direct temporal relations in the direct temporal relations corpus and the number of potential negative relations (i.e. intra-sentential temporal expression and events pairs that do not have any of the three relations).

Table 1. Relation types in the direct temporal relations dataset.

| Temporal relation type | Training set | Test set | Total |
|---|---|---|---|
| Before | 387 (17%) | 355 (20%) | 742 (18%) |
| After | 345 (15%) | 299 (16%) | 644 (16%) |
| Overlap | 1517 (68%) | 117 (64%) | 2689 (66%) |
| Total | 2248 (100%) | 1827 (100%) | 4075 (100%) |
| Potential NoRel | 2153 | 2066 | 4219 |

There were 2,248 temporal relations in the train dataset, out of which 68% were Overlap, 17% were Before, and 15% were After. There were 2,153 potential negative relations. The test dataset contained 1,827 temporal relations, with a distribution of 64% Overlap, 20% Before, and 16% After. There were 2,066 negative relations. Clearly, the dataset is highly skewed towards the Overlap relations.

## Models

### BERT

Now fairly well-known, BERT uses multiple layers of Transformer encoders to generate representations of input sequences in the output, producing a representation for each word and for the entire input sequence.[15] The principle component of the Transformer encoder is the multi-headed self-attention mechanism. Self-attention produces a word representation based on any arbitrary word positions of the input sequence by comparing each sequence member with each other sequence member (self-attention) and producing a series of probability distributions to assign importance. The multi-headed attention can simultaneously optimize for multiple such attention patterns. BERT is pre-trained using a large general domain English language corpus to predict a randomly masked word in the input (MLM – masked learning model) and to predict the next sentence (NSP – next sentence prediction).

### RoBERTa

RoBERTa is a BERT model with improved pre-training that includes: (1) training the model longer, with bigger batches, over more data (10x); (2) removing the next sentence prediction objective; (3) training on longer sequences; and (4) dynamically changing the masking pattern applied to the training data. The resulting model improved performance on several general domain NLP tasks.[19]

### ALBERT

ALBERT differs from BERT in that it incorporates two parameter reduction techniques to improve scaling of the pre-trained models.[20] First, the size of the hidden layers (in pre-training of embeddings) is separated from the size of vocabulary embedding, making it easier to grow the hidden size without increasing the parameter size of the vocabulary embeddings. Second, the model uses parameter sharing across layers preventing the parameters from growing with the depth of the network. A self-supervised loss for sentence-order prediction (SOP) prediction was used instead of the next sentence prediction (NSP) as was done in the original BERT.

### XLNet

XLNet improves upon BERT through pre-training that uses a permutation language model instead of the masked learning model (MLM) of BERT.[21] Instead of masking a token and training the model to predict it, XLNet uses different permutations of the input tokens and trains the model to predict a token using the preceding tokens in each of the permutations. In addition to the novel pretraining objective, XLNet also improves architectural designs for pretraining (See Yang et al[21] for details) resulting in a variant that has improved performance on several general domain NLP tasks.

*BioBERT*

Since biomedical domain text contains a number of domain specific terms, such as *BRCA1*, *transcriptional*, and *antimicrobial*, which appear more frequently only in the biomedical texts, BioBERT was pre-trained on MEDLINE (PubMed) abstracts and PubMed Central (PMC) full-text articles (total 18.0B words compared to the original BERT pre-training corpus of 3.3B words). It was shown to outperform the original BERT in several biomedical named entity recognition, gene-disease/chemical relation extraction, and BioASQ biomedical question answering tasks.[22]

*ClinicalBERT*

ClinicalBERT was additionally pre-trained with clinical documents. In one variant, the original BERT was the starting point, and in another BioBERT was the starting point. The clinical text from approximately 2M notes in the MIMIC-III dataset was used for ClinicalBERT pre-training. Separate pre-trained models were built using only clinical notes and only discharge summaries.[23] In presenting our results, we qualify specific variant of ClinicalBERT that we used.

For training the model, each sentence with an annotated temporal relation was prepared by adding the "<t>" and "</t>" tags as additional tokens before and after the temporal expression, and "<e>" and "</e>" tags around the event. (Models' tokenization was modified to recognize these tags as single tokens.) These tagged sentences along with theirs labels, i.e. Overlap, Before, or After, were used to create the training dataset. The potential negative samples are formed by considering all intra-sentence event and temporal expression pairs in the dataset that were not in the positive samples. Events and temporal expressions in the negative sample sentences were also tagged as in the positive samples, and the label NoRel was used. Actual positive relations and negative relations used in an experiment depended on the sampling strategy (explained later). In fine-tuning the models, the training samples (randomly shuffled) are presented one at a time in batches (as per batch size parameter).

The test dataset was prepared by considering every intra-sentential event and temporal expression pair, tagging them as in the training dataset, and using the sentences so tagged as input to a model which predicted their labels. Note that a sentence might be present more than once in training and/or in test if more than one event-temporal expression pair were present in the sentence. After the predictions were obtained, the test gold standard annotations were compared with the predictions to calculate performance metrics.

# Experiments

*Comparison of base models:* First, we compared the above described variants of BERT, i.e. the original BERT, ClinicalBERT, BioBERT, ALBERT, XLNet, and RoBERTa. As noted above, the first three models have the same model architecture and were pre-trained on different corpora while the latter three have variations in model architectures and/or training objectives. The models have a large number of parameters - 110M-125M for the base models and 340M-355M for large models. To perform a fair comparison, first we used only the base models, in part because ClinicalBERT used only the base BERT model. We used the hyperparameters shown in Table 2 in our experiments. The transformers package from the huggingface GitHub repository (*https://github.com/huggingface/transformers*, accessed February 2020) was used for the implementation of the models in our experiments. The SVM model results from Lee et al[8] were used as the baseline.

*Table 2. Hyperparameters for the models*

| | |
|---|---|
| Learning rate | 2e-6 |
| Training epochs | 20 |
| Batch size | 16 |
| Maximum sequence length | 80 |

*Sampling for training the models:* Since relation types in the training data are imbalanced (Table 1) and the potential negative samples are more than the highest frequency relation, we investigated several training data balancing strategies, including no balancing, random down sampling, and data augmented up sampling. We conducted these experiments using the RoBERTa base model. First, we used all potential negative instances ("NoRel" instances). Since the direct temporal relations are intra-sentential, the NoRel instances are also intra-sentential. In random down sampling, we randomly down sampled NoRel instances to the number of Overlap instances (the highest frequency relation). In data augmented up sampling, we used the *nlpaug* package (https://github.com/makcedward/nlpaug, accessed February 2020) to create two augmented datasets by up sampling the positive instances to the number of NoRel instances, which was 2,153. The first augmented dataset was created by randomly swapping or deleting words (the RandomWord augmenter). The second augmented dataset was created by sequentially apply four augmenters: a synonym augmenter which substitutes similar word according to the WordNet synonyms; a word embedding augmenter which inserts or substitutes words according to the *fasttext* word embeddings; a contextual word embedding augmenter which inserts or substitutes words using the BERT pre-trained representations; and the RandomWord augmenter.

*Model size impact:* Lastly, we investigated whether the model size has an impact on the performance of direct temporal relation classification. We first compared the RoBERTa base model with the RoBERTa large model using the random down sampling strategy. The based model had 12-layers, 768-hidden units, 12-heads, and 125M total parameters and the large model had 24-layers, 1024-hidden units, 16-heads, and 355M total parameters. A larger model typically benefits from larger training datasets, so we also compared the base and large models using the four augmenter up-sampling.

## Results

The base model results for the direct temporal relations are shown in Table 3 (with NoRel instances randomly down sampled to the Overlap frequency). The Table shows performance for the original BERT model, ALBERT (which is only available in large model), BioBERT, three variants of ClinicalBERT, XLNet, and the RoBERTa base model. Included is the baseline performance (0.637 F measure) of a feature-engineered SVM model reported in Lee et al.[8] Interestingly, the original BERT did not improve performance over SVM and in fact a distinct loss of recall can be seen. However, other models steadily improved performance over SVM. ClinicalBERT pretrained on BERT with 150,000 clinical notes achieved the highest precision (0.741) whereas ClinicalBERT pretrained on BioBERT with 100,000 discharge summaries achieved the highest recall (0.693). XLNet and RoBERTa achieved the highest F measure, which was 0.705. It should be noted that clinical NER extraction studies showed a smaller improvement with RoBERTa and ClinicalBERT.[24] The improvement achieved by XLNet and RoBERTa is 0.068 F measure (on the 1.000 scale) in absolute terms and a 18.73% error reduction rate relative to the SVM performance.

*Table 3. Comparison of base models for direct temporal relation extraction. Precision (P), Recall (R), and F measure (F) are shown.*

| Model | P | R | F |
|---|---|---|---|
| SVM (Lee et al) | 0.639 | 0.636 | 0.637 |
| BERT | 0.676 | 0.585 | 0.627 |
| ALBERT (large) | 0.664 | 0.648 | 0.655 |
| BioBERT | 0.731 | 0.607 | 0.663 |
| ClinicalBERT (BioBERT_150000) | 0.678 | 0.690 | 0.684 |
| ClinicalBERT (BERT_150000) | **0.741** | 0.642 | 0.688 |
| ClinicalBERT (BioBERT_100000) | 0.696 | **0.693** | 0.695 |
| ClinicalBERT (BERT_100000) | 0.730 | 0.667 | 0.697 |
| XLNet | 0.719 | 0.691 | 0.705 |
| RoBERTa (base) | 0.734 | 0.678 | **0.705** |

*Table 4. Performance comparison of SVM and RoBERTa by Temporal Relation Type*

| Relation Type | Distribution in the test set | SVM | | | RoBERTa | | |
|---|---|---|---|---|---|---|---|
| | | P | R | F | P | R | F |
| After | 16% | **0.534** | 0.338 | 0.314 | 0.498 | **0.580** | **0.536** |
| Before | 20% | 0.569 | 0.422 | 0.485 | **0.581** | **0.581** | **0.581** |
| Overlap | 64% | 0.667 | **0.777** | 0.718 | **0.841** | 0.721 | **0.776** |
| Total | 100% | 0.639 | 0.636 | 0.637 | **0.734** | **0.678** | **0.705** |

A detailed per-relation type performance comparison of RoBERTa and SVM is shown in Table 4. SVM had higher precision for the After relation (the lowest frequency relation) and higher recall for the Overlap relation (the highest frequency relation). But, for all other cases including the F measures for all three relation types, RoBERTa achieved distinctly better performance.

*Table 5. Impact of various data sampling methods on RoBERTa performance*

| Training data sampling method | P | R | F |
|---|---|---|---|
| No sampling | 0.687 | 0.700 | 0.693 |
| RandomWord up-sample positives to NoRel | 0.670 | **0.739** | 0.703 |
| Four augmenters up-sample positives to NoRel | 0.728 | 0.681 | 0.704 |
| Random down-sample of NoRel to Overlap | **0.734** | 0.678 | **0.705** |

Different sampling methods for training were compared in Table 5 using the RoBERTa base model. No sampling, i.e. using the natural distribution of three relation types and all potential NoRel instances in the training data, is the baseline here. Data augmented up-sampling and random down sampling perform nearly similar in terms of the F measure, and better than the no-sampling. We postulate that this low variation may be due to already high performance of RoBERTa, coupled with relatively small number of potential negative samples. However, RandomWord up-sampling has distinctly high recall while the random down sample has the highest precision. Due to the simplicity of random down sampling, we consider it as the winner.

We compared performance of the large and base models, 80 and 128 input maximum sequence lengths, and 16 and 32 training batch sizes in Table 6 (using RoBERTa). It should be noted that, in the training data, 94 percentile sequence length was 80 and the maximum sequence length was 128. The large model achieved better performance than the base model, and the smaller batch size performed better. While the model is sensitive to the input maximum sequence length, RoBERTa didn't benefit from increasing the sequence length to 128.

*Table 6. Model size comparison*

| Model setting | P | R | F |
|---|---|---|---|
| RoBERTa base 80/16 | 0.7425 | 0.6264 | 0.6795 |
| RoBERTa large 80/16 | 0.7578 | **0.6932** | **0.7241** |
| RoBERTa base 80/32 | 0.7293 | 0.6449 | 0.6845 |
| RoBERTa large 80/32 | **0.7732** | 0.6428 | 0.7020 |
| RoBERTa base 128/16 | 0.7340 | 0.6780 | 0.7050 |

In summary, the large RoBERTa model achieved the highest F measure (with 80-word sequence length and 16 batch size), 0.7241, and the highest recall, 0.6932, with a moderately high precision. This best performance is 0.0871 improvement in F measure (on the 1.000 scale) in absolute terms, and a significant 24% error rate reduction relative to the SVM model.

## Discussion

Temporal relations are intrinsically difficult to extract accurately because of subtle differences in the way temporal expressions can be written and how they are related to events in the text. For example, in the sentence from the dataset, "_Several days prior to discharge_, the patient developed _some erythematous rash_ under her …", the temporal expression has the complexity of stating an event (underlined) that is shifted in time from another event ("discharge") thus changing the temporality of the relationship. In the light of such complexity, the 2012 i2b2 temporal relations challenge, with relations defined between events and events, temporal expressions and temporal expressions, and events and temporal expressions, was an ambitious undertaking. Initially eight relation types were proposed but later due to low inter-annotator agreement they were reduced to the three relation types studied here. Even then the inter-annotator agreement was low (0.3 kappa, for approximate entity spans). We believe that several distinct NLP tasks may have been unintentionally combined into one in the challenge.

The direct temporal relations dataset Lee et al constructed was an equally challenging dataset but was focused on a specific task. Since the subset was further reviewed by medical experts to ensure that it met the simplified semantic meaning, it formed a sound basis for research in temporal relation extraction.

Our best model produced new state-of-the-art performance on the direct temporal relations dataset by achieving 0.7241 F measure and reducing error rate by 24% relative to the previous best. While direct comparison is not possible, we show in Table 7, how the performance of our best model, the SVM model of Lee et al, and the systems built for the original 2012 i2b2 temporal relations challenge compare. Note that the original challenge systems have low F measures, although the Vanderbilt system had the highest recall (accompanied by very low precision).

_Table 7. Comparison with other studies_

| Dataset | System | P | R | F |
|---|---|---|---|---|
| Original 2012 i2b2 temporal relations | Original Vanderbilt system | 43.53 | **76.99** | 55.61 |
| | Re-trained Vanderbilt system | 64.16 | 49.15 | 55.66 |
| | Syntactic graph kernel based system | 64.46 | 54.27 | 58.92 |
| | CRF-based system | 48.51 | 39.52 | 43.56 |
| | | | | |
| Direct temporal relations | SVM-based system | 63.93 | 63.62 | 63.77 |
| | Roberta base system (ours) | 73.40 | 67.80 | 70.50 |
| | Roberta large system (ours) | **75.78** | 69.32 | **72.41** |
| | | | | |
| THYME dataset "contain" relation | BioBERT (PMC) - TS | 67.30 | 69.50 | 68.40 |

Another point of comparison is the results from a recent study that analyzed the *contain* temporal relation on the THYME corpus (which was also used in SemEval-2017 Task 12). Unfortunately, the dataset is not available for public distribution at this time and so we could not make a direct comparison. The contain relation indicates that an event occurred entirely within the temporal bounds of a narrative container. The study considered event-to-time, time-to-event, and event-to-event relations, and the best performance was achieved using BioBERT. We showed these results also in Table 7. While not directly comparable, RoBERTa in our study achieved 0.7241 F measure, which is significantly higher than 0.684 F measure reported in the THYME-data study for the contain relation.

Why does RoBERTa perform better on temporal relations? The attention-based contextual language models are known to perform well on most NLP tasks. A unique contribution of our study is to compare how different variants of these models perform on clinical temporal relations. Specifically, we show that pre-training with a large corpus as in RoBERTa/XLNet or training with clinical and biomedical corpus as in ClinicalBERT performs better than SVM or the original BERT in temporal relation extraction. We believe that a better and targeted representation of words in these models contributes to the performance difference.

Informal error analysis revealed many challenges with temporal annotations and temporal information extraction, as was also observed in previous research.[13,25] We found that often system predictions of Overlap were Before or After in the gold standard. Further analysis indicated the possibility of annotation errors. For example, *"<e> The patient's blood pressure </e> remained stable on <t> hospital day #2 </t> through hospital day #3 status…"* was annotated as After but the system (in our view) correctly predicted it as Overlap.

In some cases, inconsistencies in event annotation may have further complicated temporal reasoning. For example, in the two event annotations *"prior to discharge <t> today </t>…"* and *"<t> several days prior to discharge </t>…"* the word "prior" was treated differently relative to the nearby event annotation. While it is arguable how to best annotate a temporal event in such cases, the choice of annotation becomes a critical factor in temporal reasoning.

Certain limitations of our work are worth noting. First, the implication of using the direct relations dataset is that the relations are limited to within a sentence. Second, we have only analyzed relations between events and temporal expressions. Lastly, while we experimented with several BERT variants, the list is by no means exhaustive. In the future work, we hope to explore temporal relations beyond event-temporal expressions, and study the contrast between models that are extensively pre-trained (e.g. RoBERTa) and models that are pre-trained on domain specific corpus.

## Conclusion

In this study we established new state-of-the-art performance for clinical temporal relations using RoBERTa, a BERT model that was trained on a 10x larger corpus compared to the original BERT. We demonstrated the temporal relations extraction performance using a semantically focused subset of the 2012 i2b2 Challenge temporal relations, called direct temporal relations. The new dataset is not only simpler but also has a semantically coherent theme where a (direct) temporal relation is defined as a temporal expression modifying an event or vice versa, or the time expression and the event being arguments of the same predicate in a parse tree.

We compared variously trained BERT models and architectural variants, and showed that BERT trained on clinical text is better than the original BERT, ALBERT, and BioBERT for this task. We also observed that XLNet and RoBERTa, which use different learning strategies on general domain corpus, have outperformed the BERT trained on clinical text (ClinicalBERT). Further studies are needed to understand the tradeoffs between domain-customization and better training strategies.

Our study also observed that simple random down-sampling of negatives is good enough for training the model for the dataset, however, we also note that the number of potential negative samples were not significantly more than the highest frequency relation due to the intra-sentential nature of the direct temporal relations. We also found that the larger models tend to perform better than the base models at the cost of long running times and GPU memory requirements. In summary, this study showed that RoBERTa large model achieves 0.0871 absolute improvement in F measure (on the 1.00 scale) and an impressive 24% error rate reduction relative to the F measure achieved by the previous state-of-the-art model (an SVM).

## Conflict of Interest

Dr. Xu and The University of Texas Health Science Center at Houston have research-related financial interests in Melax Technologies, Inc.